# A Novel Trend Symbolic Aggregate Approximation for Time Series

Yufeng Yu, Yuelong Zhu, Dingsheng Wan,Qun Zhao
College of Computer and Information, Hohai University, China
+8613951670067
hhuheiyun@126.com

Huan Liu
Computer Science and Engineering,
Arizona State University,U.S.A

## ABSTRACT
Symbolic Aggregate approximation (SAX) is a classical symbolic approach in many time series data mining applications. However, SAX only reflects the segment mean value feature and misses important information in a segment, namely the trend of the value change in the segment. Such a miss may cause a wrong classification in some cases, since the SAX representation cannot distinguish different time series with similar average values but different trends. In this paper, we present Trend Feature Symbolic Aggregate approximation (TFSAX) to solve this problem. First, we utilize Piecewise Aggregate Approximation (PAA) approach to reduce dimensionality and discretize the mean value of each segment by SAX. Second, extract trend feature in each segment by using trend distance factor and trend shape factor. Then, design multi-resolution symbolic mapping rules to discretize trend information into symbols. We also propose a modified distance measure by integrating the SAX distance with a weighted trend distance. We show that our distance measure has a tighter lower bound to the Euclidean distance than that of the original SAX. The experimental results on diverse time series data sets demonstrate that our proposed representation significantly outperforms the original SAX representation and an improved SAX representation for classification.

## 1 INTRODUCTION
Time series is a sequence of data changing with time order, which is increasingly important and has attracted an increasing interest due to its wide applications in many domains, such as nature science, engineering technology and social economics. Therefore, it is said that the time series mining is considered as one of the ten challenging problems in data mining [1]. However, there are a number of challenges in time series data mining, such as high dimensionality, high volumes, high feature correlation and large amount of noises. Moreover, most time series data exist as collections of consecutive values varying continuously in time, which makes many data mining methods ineffective and fragile. As a prerequisite, the consecutive time series data need to be taken a way of dimensionality reduction and then formed new representations for time series data mining algorithms.

Symbolic Aggregate approximation (SAX) [2] is a classical symbolic approach for time series data mining, the basic concept of which is to convert the numerical form of a time series into a sequence of discrete symbols according to designated mapping rules. SAX can reduce dimensionality/ numerosity of data and has a lower bound to the Euclidean distance, that is, the error between the distance in the SAX representation and the Euclidean distance in the original data is bounded [3]. Therefore, the SAX representation speeds up the data mining process of time series data while maintaining the quality of the mining results. The SAX has been widely used for applications in various domains such as mobile data management [4], financial investment [5] and shape discovery [6].

SAX is based on Piecewise Aggregate Approximation (PAA) [7], so, it has a major limitation inherit from PAA on dimensionality reduction. That is, the symbols in the SAX representation are mapped from the mean values of segments, which may miss other important feature information such as trend and extreme feature of the segments. Furthermore, different segments with similar average values may be mapped to the same symbols, which make the SAX distance between them is 0. For instance, two time series (a) and (b) in Fig.1 have different trend and extreme feature but their SAX representations are the same as *'feacdb'*, which may cause misclassifications when using distance- based classifiers.

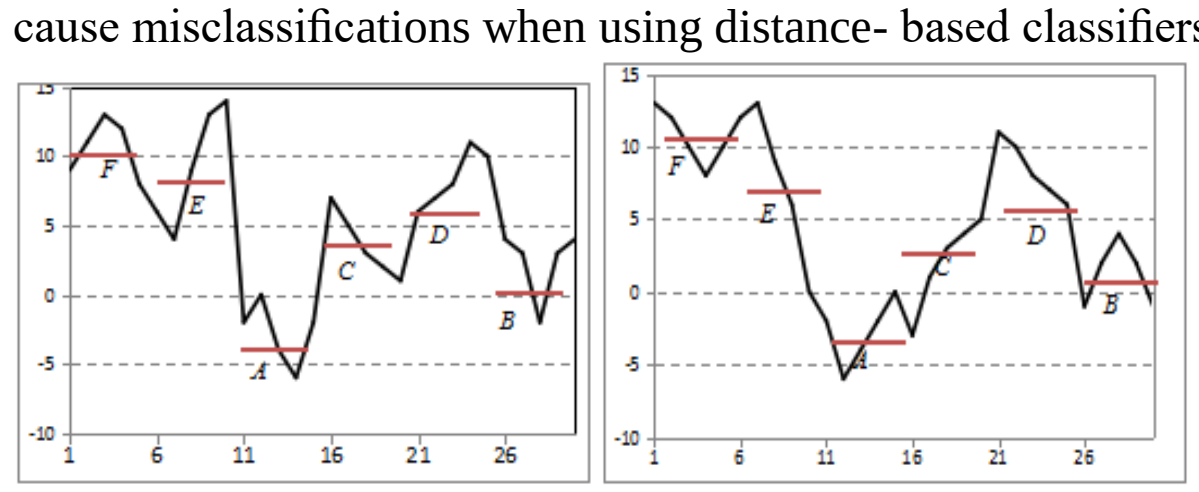

(a) (b)

**Figure 1: same SAX representation in same condition**

Fig.1. (a) and (b) have the same SAX symbolic representation '*FEACDB'* in the same condition where the length of time series is 30, the number of segments is 6 and the size of symbols is 6. However, they have different time series. (a) Time series 1, (b) Time series 2.

The Extend SAX (ESAX) representation overcomes some limitations by tripling the dimensions of the original SAX [8], which adds two new values in each segment based on SAX, i.e., max value and min value, respectively. Therefore, the ESAX representations of time series (a) and (b) in Fig.1 are '*efffecaaaacffdbcbb'* and '*effcefbaafcaadfbbc'* respectively. Though ESAX can more approximately represent the time series, it still has the problem in terms of ignoring the important trend feature.

We present a novel symbolic aggregate approximation representation and distance measure for time series, which can not only reflect the segment mean value feature, but also capture trend feature with good resolution, further support time series data mining tasks.

Our work can be simply summarized as follows:

•We extract the trend feature from time series, according to trends distance factor and trend shape factor. Then we design multi-resolution discretization method to transform the trend feature into symbols.

•We propose a novel symbolic dimensionality reduction approach, and call it as Trend Feature Symbolic Aggregate approximation(*TFSAX*) and put forward distance function *TDIST* based on *TFSAX*.

•We demonstrate the effectiveness of the proposed approach by experiments on different datasets. The comprehensive experiments have been conducted in comparison with the SAX and the ESAX representations, experiments validate the utility of our proposed approach.

## 2 BACKGROUND

SAX[2] is a symbolic aggregation approximation representation method based on piecewise aggregation approximation PAA [7]. For instance, to convert a time series sequence $C$ of length n, $C=\{C_1,C_2,\ldots,C_n\}$, into w symbols, the SAX works as follows:

Step 1: Normalization. Transform raw time series C into normalized time series C′ with mean of 0 and standard deviation of 1.

Step 2: Dimensionality reduction. The time series is divided into $w$ equal-sized segments by Piecewise Aggregate Approximation (*PAA*) [7]. That is, $\bar{C}=\{\bar{C}_1, \bar{C}_2\ldots \bar{C}_w\}$,the $i^{th}$ element of $\bar{C}_i$ is the average of the $i^{th}$ segment and is calculated by the following equation:

$$\bar{C}_i=\frac{w}{n}\sum_{j=\frac{n}{w}(i-1)+1}^{\frac{n}{w}i} c_j \qquad (1)$$

Where $c_j$ is one point of time series $C$, $j$ is from the starting point to the ending point for each segment.

Step 3: Discretization. According to *SAX* breakpoints search table, choose alphabet cardinality, discretize $\bar{C}$ into symbols and obtain SAX representation $\hat{C}$ .

Breakpoints are a sorted list of numbers $B=\beta_1,\beta_2,\ldots,\beta_a$ such that the area under a $N(0,1)$ Gaussian curve from $\beta_i$ to $\beta_{i+1}=\alpha$, where $\beta_1=-\infty$, $\beta_\alpha=+\infty$. Comparing the segment mean value of $\bar{C}$ with breakpoints, if the segment mean value is smaller than the smallest breakpoint $\beta_1$, the segment is mapped to symbol ‘a’; if the segment mean value is larger than the smallest breakpoint $\beta_1$ and smaller than$\beta_2$, the segment is mapped to symbol ‘b’; and so forth. These symbols for approximately representing a time series are called a ‘word’. Fig. 2 illustrates a sample time series converted into the SAX word representation.

**Table 1 A Lookup Table for Breakpoints**

| $\beta_i$ \ $a$ | 3 | 4 | 5 | 6 | 7 |
|---|---|---|---|---|---|
| $\beta_1$ | -0.43 | -0.67 | -0.84 | -0.97 | -1.07 |
| $\beta_2$ | 0.43 | 0 | -0.25 | -0.43 | -0.57 |
| $\beta_3$ | - | 0.67 | 0.25 | 0 | -0.18 |
| $\beta_4$ | - | - | 0.84 | 0.43 | 0.18 |
| $\beta_5$ | - | - | - | 0.97 | 0.57 |
| $\beta_6$ | - | - | - | - | 1.07 |

For the utilization of the SAX in classic data mining tasks, the distance measure was proposed. Give two raw time series $Q=\{q_1, q_2,\ldots, q_n\}$ and $C=\{c_1, c_2,\ldots, c_n\}$ with the same length n, $\hat{Q}$ and $\hat{C}$ are their SAX representations respectively with the word size $w$. In order to measure the similarity based on SAX representation, the SAX distance MINDIST is defined as follows:

$$MINDIST(Q,C)=\sqrt{\frac{n}{w}}\sqrt{\sum_{i=1}^{w}(dist(q_i,\hat{c}_i))^2} \qquad (2)$$

where *dist()* function can be implemented using the lookup table as illustrated in Table 1, and is calculated by the following equation:

$$dist(q_i,\hat{c}_i)=\begin{cases}0 & |q_i-\hat{c}_i|\le 1\\ \beta_{max(q_i,\hat{c}_i)-1}-\beta_{min(q_i,\hat{c}_i)} & otherwise\end{cases} \qquad (3)$$

Since SAX extracts the mean value information, and misses the trend information, Consequently, MINDIST only measures the similarity of mean value of segments through dimensionality reduction; it can’t evaluate the trend similarity.

## 3 RELATED WORK

SAX is the first proposed symbolic approach which allows dimensionality reduction and supports lower bounding distance measure. Though SAX has a generality of the original presentation and works well in many problems, it can also lead to the loss of trend feature information within the original sequence. There have been some improvements of the SAX representation recently.

Some methods improve the SAX by adaptively choosing the segments. The method in [9] uses the discretization of Adaptive Piecewise Constant Approximation (APCA) [10] to replace the PAA[7] in the SAX. The method in [11] makes use of an adaptive symbolic representation with the adaptive vector of ‘breakpoints’. While the two methods above reduce the reconstruction error on some data sets, they still use the average values as the basis for approximation (the latter method uses the same distance measure as the SAX) and do not consider the differences of value changes between segments.

Some methods improve the SAX by enriching the representation of each segment. The method in [8] uses three symbols, instead of a single symbol, to represent a segment in time series. This method triples the dimensions of the SAX and the high dimensionality increases the computational complexity. The method in [12] utilizes a symbolic representation based on the summary of statistics of segments. The method considers the symbols as a vector, including the discretized mean and variance values as two elements. However, it is may be inappropriate to transform the variances to symbols using the same breakpoints for the transformation of the mean values to symbols.

Trend estimation of time series is an important research direction. Many methods have been proposed to represent and measure trends[13-22].Sun[19] proposed Symbolic Aggregate approx−imation based on Trend Distance (SAX-TD), which captures the trends of time series in numerical form by approximating the measure of trends using the difference between the average and the starting point of segment, and the difference between the average and the ending point of segment. Though SAX-TD can better represent time series with different trend characteristics, it is difficult to reflect the trend of the segmentation when the starting point and ending point of the sequence are equal to the mean. Yin [22] proposed an improved SAX method called Trend Feature Symbolic Approximation (TFSA), which uses the trend symbols to represent the subsequences after segmentation and allows the subsequences can visually display these trend features. Though TFSA represents biomedical series using trend symbols to improve the classification accuracy of symbolic methods, it also triples the dimensions of the SAX and the high dimensionality increases the computational complexity.

# 4 TFSAX-IMPROVED SAX BASED ON TREND FEATURE

As we reviewed above, SAX maps the time series segments to symbols by their average values. This representation is imprecise when the trends of the segments are different but with similar average values. Although there are no common definition and a measurement of trend in time series, the distance factor and internal shape factor are important in segment trend estimation. The former measure the first-order difference between the starting and the ending points of the series, it can quantitatively measure the trend distance rather than qualitatively define such as 'significant up' and 'significant down' of the time series. However, the distance factor can only represent the overall trend; it may not indicate the partial trend of the sequence. For example, there may be partial increasing and decreasing trends in the overall stationary time series, and there may also be partial decreasing and stationary trends in the overall increasing time series. The internal shape factor records the variation of the trend of the time series, which may play an important role in short-term trend research of time series.

In this paper, we employ the sequence mean feature and trend shape feature to represent the time series. It first constructions trends feature triangular shape using trend distance factor and trend shape factor as its right-angle edge, and then, adopt tangent function value of the angle θ which calculate as the ratio of trends distance factor and internal shape factor to quantitatively measure the different trends, at last, incorporate the trend feature variations into the original representation. This method represents the time series using both average feature and trend morphological feature, it overcomes the shortcomings of the traditional symbolic method that only uses the mean values to describe the original time series. The experiments' proved that this method can better represent different forms of time series and can improve the efficiency and accuracy of time series symbolic represent.

## 4.1 Trend Feature Extraction

*4.1.1 Trend distance factor*

Trends are important characteristics of time series, and they are crucial for the analysis of similarity and classification of time series [24].However, there is no common definition and a measurement of trend in time series. Therefore, how to define the measure of the distance factor between trend series becomes a difficult problem need to be solved while mining a time series by using trend characters.

Although there are no common definition of trend and a measurement of trend distance in time series, the starting and the ending points are important in segment trend estimation. For example, a trend is up when the value of the ending point is significantly larger than the value of the starting point, while the trend is down when the value of the ending point is significantly smaller than the value of the starting point. It is difficult to qualitatively define a trend, such as the definitions of 'significant up' and 'significant down', 'significant down' and 'slight down'. However, if the trend information of a segment is not utilized, the representations of a time series containing many segments are rough.

**Definition 1.Trend distance:** Given time series segments $q$ and $c$ with the same length, defined $\Delta q(t_s)$ and $\Delta q(t_e)$ to represent the first order difference between the starting point, the end points and the sequence mean of the segment q separately. The trend distance factor $td$ of them is defined as follows:

$$td = \Delta q(t_s) - \Delta q(t_e) \qquad (4)$$

More obviously, it can see that sequences (a) and (b) in Fig.2 have the same trend(both rising), but which can be distinguished by the trend distance factor $\Delta p(t_s)$-$\Delta p(t_e)$.

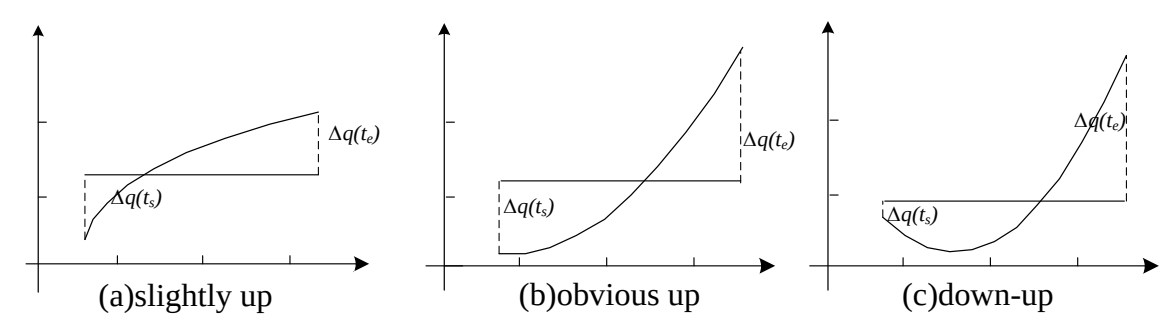

**Figure 2: Several typical segments with the same trend**

*4.1.2 Trend shape factor*

For a time series sequence, the start and end points of the sequence can determined the overall trend, but cannot fully indicate the trend shape features in the sequence. For example, there may be local up-trend and in the overall stationary time series, similarly to an overall up-trend time series, which may contain local down-trend and stationary sequence. So, it is necessary to use the feature point such as extreme point or trend point to measure the long time series sequence.

The trend point is the turning point of the time series sequence from one shape to another, which is the point of change that affects the local trend of the pattern.

**Definition 2. Trend point:** Given a time series $X$ of length N, $X=\{x_1,x_2,\ldots,x_n\}\in R^N$ and a data point $x_i$ (i<m), if $x_i$ satisfied 1) $(x_i-x_{i-1})(x_{i+1}-x_i)<0$; 2)$(x_i-x_{i-1})(x_{i+1}-x_i)=0$ and $(x_i-x_{i-1})\neq(x_{i+1}-x_i)$ $(2\leq i\leq n)$, then $x_i$ is called a up (down, stationary)-trend point.

Correspondingly, the trend shape factor of the time series can be transformed into find the trend point numbers $K$ of the time series. The larger the K takes, the more complex the trend be expressed. Particularly, the value of $K$ is 1 for overall up-, down or stationary trend sequence.

Fig.3 lists several typical sequences which have same (or similar) overall trend but different trend shape features. These sequences can be divided into different trend shape, overall up shape, up after down shape (down-up), down after up shape and compound shape which may contains several combinations of up-down-up shapes.

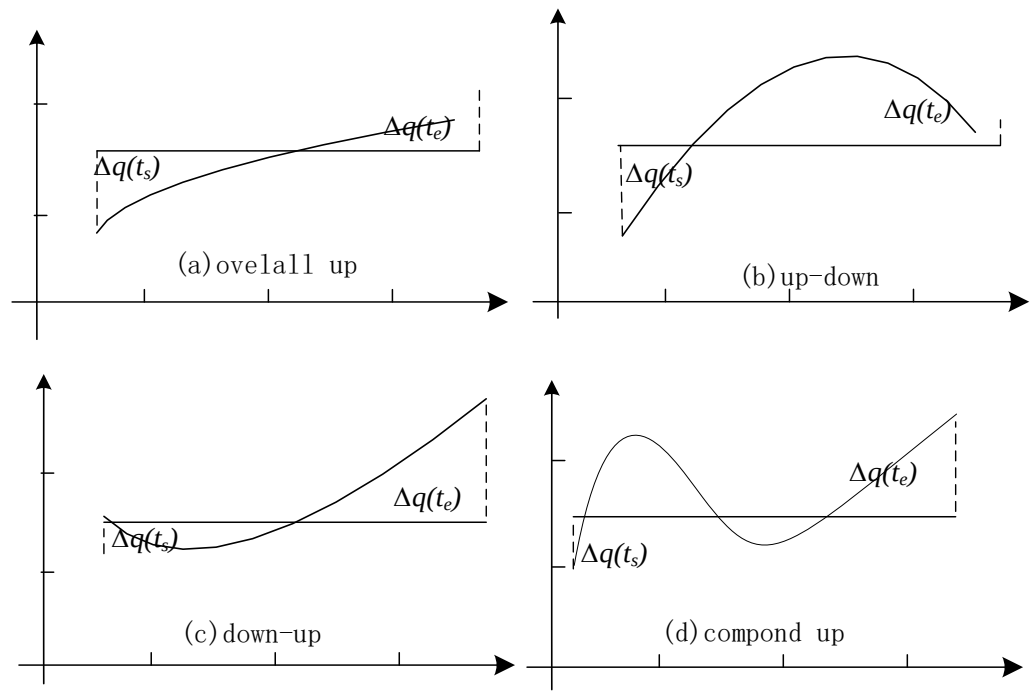

**Figure 3: Several typical patterns with the same (or similar) overall trend but different trend shape**

*4.1.3 Trend Feature Representation*

As we reviewed above, only used trend distance factor or the trend shape factor cannot completely represent the feature of the trend. Therefore, this paper proposes a new approach to represent the overall feature. It first constructions trends feature triangle using trend distance factor ($\Delta p(t_e)$-$\Delta p(t_s)$ )and trend shape factor ($K$) as its right-angle edge, and then, adopt tangent function value of the angle $\theta$ which calculate as the ratio of trend distance factor and trend shape factor to quantitatively measure the different trends, that is:

$$tan\theta = \frac{\Delta p(t_e) - \Delta p(t_s)}{N} \tag{5}$$

The construction of trend feature triangle is shown in Fig. 4.

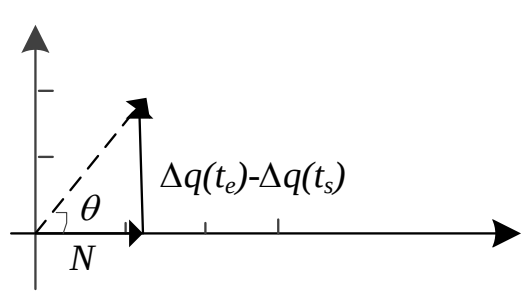

**Figure 4: The construction of trend feature triangle**

The angle space corresponding to the trend feature triangle is (-90°, 90°), which can be divided into certain numbers of no overlapping intervals, each interval corresponds to a symbol. The trends of segment can be catalogued into three kinds based on the value the angle $\theta$ taken. The $\theta$ takes a positive value, it indicates that the sequence has the up-trend, moreover, the higher value the $\theta$ takes, the faster the trend rising (slow rising, fast rising and sharp rising); so as to the stationary trend and the down-trend. Based on the extent of trend change from judging whether it is high or low and whether it is slow or sharp, design multi-resolution angle break-point intervals search table to map trends to symbols, which is shown as Table 2

**Definition 3. Angle Breakpoints (AB)**: Angle breakpoints are sorted list of angles, denoted by $B=\theta_1\theta_2\ldots\theta_\alpha$. Suppose each interval from $\theta_i$ to $\theta_{i+1}$ has the equal probability of $1/\alpha$, specifically, $\theta_1 = -\infty, \theta_\alpha = +\infty$.

**Table 2 Angle Breakpoint Interval SearchTable**

| $\theta_i$ \ $a$ | 2 | 3 | 4 | 5 | 6 |
|---|---|---|---|---|---|
| $\theta_1$ | 0 | -5° | -30° | -30° | -30° |
| $\theta_2$ | | 5° | 0° | -5° | -5° |
| $\theta_3$ | | | 30° | 5° | 0° |
| $\theta_4$ | | | | 30° | 5° |
| $\theta_5$ | | | | | 30° |

The corresponding relationship between trend symbol and angle is determined by search table shown as Table 2, for each slope value in the segment, use symbol to represent trend feature according to Table 2. Comparing the angle reflected by slope with a series of angle breakpoints, if the angle of trend segment is smaller than the smallest angle breakpoint $\theta_1$, this trend segment is mapped to symbol *'A'*; if the angle of trend segment is larger than the smallest angle breakpoint $\theta_1$ and smaller than $\theta_2$, this trend segment is mapped to symbol *'B'*; and so on.

*4.1.4 Trend Feature Distance*

Distance measurement is an important problem in data mining. So it is necessary to define the trend feature distance after the trend feature representation.

**Definition 4: Trend feature distance (*tfdist*):** The distance between trend characters can be described by a two-dimensional matrix, the element *tfdist*[i][j] corresponding to the $i^{th}$ row, the $j^{th}$ column can be calculated as follow:

$$tfdist[i][j] = \begin{cases} 0 & |i - j| \le 1 \\ tan(\theta_{max(i,j)-1} - \theta_{min(i,j)}) & otherwise \end{cases} \tag{6}$$

By checking the corresponding row and column of two symbols in Table 3, the distance between trend symbols can be obtained. For example, *tfdist (A,B)=0, tfdist (A,C) =tan25°, tfdist (A,E)= tan60°*. When trend cardinality $\alpha$ is arbitrary, the cell′ in *tfdist* () search table can be calculated by Equation (6).

**Table 3** *tfdist* () Search Table ($\alpha$=5)

| trend character | trend character | | | | |
|---|---|---|---|---|---|
| | *A* | *B* | *C* | *D* | *E* |
| *A* | 0 | 0 | *tan25* | *tan35* | *tan60* |
| *B* | 0 | 0 | 0 | *tan5°* | *tan35* |
| *C* | *tan25* | 0 | 0 | 0 | *tan25* |
| *D* | *tan35* | *tan5°* | 0 | 0 | 0 |
| *E* | *tan60* | *tan35* | *tan25* | 0 | 0 |

## 4.2 TFSAX Representation

*4.2.1 Symbolic Representation*

Trend Features based SAX (TFSAX) is a symbolic representation of the mean and trend features of the time series, which not only inherited the advantages of the classic SAX, but also overcomes the shortcomings of the traditional symbolic method that only uses the mean values to describe the original time series, so that it can better represent a different form of time series. Because the continuity of time series data, the ending point of a segment is the starting point of the following segment. Therefore, we only need to add a trend feature to each of the sequence segments to indicate the trend of the segment.

Formally, given two time series Q and C with the length of $n$, the representations with $w(w<<n)$ words of them are:

$$Q: \Delta q(1) q_1 \Delta q(2) q_1 ... \Delta q(w) q_w$$
$$C: \Delta c(1) \hat{c}_1 \Delta c(2) \hat{c}_2 ... \Delta c(w) \hat{c}_w$$

where $\hat{c}_1 \hat{c}_2 ... \hat{c}_w$ are the symbolic representations by the SAX, $\Delta c(1) \Delta c(2) ... \Delta c(w)$ are the trend feature representations, and so as to $q_1 q_2 ... q_w$ and $\Delta q(1) \Delta q(2) ... \Delta q(w)$ .Compared to the original SAX, our representation adds w dimensions for trend feature.

Based on the two time series shown in Fig.1, the differences between the several methods under the same conditions are specifically analyzed. The results of the SAX method are "*FEACDB*" and the results of the ESAX are "*EFFFECAAAACFFDBCBB*" and "*EFFCEFBAAFCAADFBBC*". The results of my method are "$F_dE_fA_aC_bD_eB_c$" and "$F_cE_aA_e C_fD_bB_d$". For the same number of segments, the dimension of ESAX is three times that of SAX, and the dimension of this method is approximately twice that of SAX. However, the method of my method is much higher than SAX and ESAX in improving the accuracy of distance calculation.

*4.2.2 Distance measures*

After the time series are symbolized, it is generally necessary to define distance metrics between the sequences as a metric for subsequent mining. However, for the hydrological time series data mining, the SAX method ignores the trend characteristics of the sequence and discrete the partial precision of the sequence when the time series is discretized. On the other hand, because the trend distance can reflect the trend information characteristics of the sequence segment, and can indicate the trend difference of the sequence segment. Therefore, this section combines the SAX symbol distance and the trend feature symbol distance, and proposes a time series based on the trend feature. The approximate approximation method is used to solve the distance measurement after time series data symbolization. Therefore, we define the distance between two time series based on the trend distance as follows:

$$TDIST(Q,C) = \sqrt{\frac{n}{w}} \sqrt{\sum_{i=1}^{w} (dist(q_i, \hat{c}_i)^2 + \frac{w}{n}(tfdist(q_i, c_i))^2)} \quad (7)$$

where $dist(q_i, \hat{c}_i)$ is the symbolic distance of the segments $q_i$ and $c_i$ based on SAX representation and $tfdist(q_i,c_i)$ is the trend symbolic distance of the segments $q_i$ and $c_i$ based on TFSAX representation. Note that $q_i$ and $\hat{c}_i$ are the new representations of time series $Q$ and $C$ with the same length $n$, $w$ is the number of segments (or words), and $q_i$ and $\hat{c}_i$ are the symbolic representations of segments of segments $q_i$ and $c_i$, respectively.

From Eq. (6), we see that the influence of the trend distance on the overall distance is weighted by the ratio of dimensionality reduction w/n. w/n is larger when there are more divided segments and each segment is shorter. w/n is smaller when there are fewer divided segments and each segment is longer. This is because in a short segment, the trend is likely to be linear and can be largely captured by two points and hence the weight for the trend distance is high. When the segment is long, the trend is complex, two points are unlikely to capture the trend and hence the weight of the trend distance is low.

*4.2.3 Lower bound*

One of the most important characteristics of the SAX is that it provides a lower bounding distance measure. Lower bound is very useful for controlling errors and speeding up the computation. Below, we will show that our proposed distance also lower bounds the Euclidean distance.

The quality of a lower bounding distance is usually measured by the tightness of lower bounding (*TLB*).

$$TLB = \frac{Lower\ Bounding\ Distance(P,Q)}{Euclidean\ Distance(P,Q)} \quad (8)$$

The value of *TLB* is in the range [0, 1]. The larger the TLB value, the better the quality. Recall the distance measure in Eq. (7), we can obtain that TLB(TDIST)≥TLB(MINIDIST), which means the TFSAX distance has a tighter lower bound than the original SAX distance. In conclusion, our improved TFSAX not only holds the lower bounding property of the original SAX, but also achieves a tighter lower bound.

According to [3,5], the authors have proved that *SAX* distance lower bounds the *PAA* distance and the *PAA* distance lower bounds the Euclidean distance, that is:

$$\sqrt{\sum_{i=1}^{n} (q_i - c_i)^2} \geq \sqrt{\frac{n}{w}} \sqrt{\sum_{i=1}^{w} (\bar{q}_i - \bar{c}_i)^2} \quad (9)$$

$$n(\bar{p} - \bar{q})^2 \geq n(dist(p,q))^2 \quad (10)$$

For proving the TDIST also lower bounds the Euclidean distance, we repeat some of the proofs here. Let $\bar{Q}$ and $\bar{C}$ be the means of time series $Q$ and $C$ respectively. We first consider only the single-frame case (i.e. $w$=1), Ineq.(9) can be rewritten as follows:

$$\sum_{i=1}^{n} (q_i - c_i)^2 \geq n(\bar{Q} - \bar{C})^2 \quad (11)$$

Recall that $\bar{Q}$ is the mean of the time series, so $p_i$ can be represented in terms of $q_i = \bar{Q} - \triangle q_i$. The same applies to each point $c_i$ in C. Thus, Ineq.(11) can be rewritten as follows:

$$\sum_{i=1}^{n} ((\bar{Q} - \Delta q_i) - (\bar{C} - \Delta c_i))^2 \geq n(\bar{Q} - \bar{C})^2 \quad (12)$$

Rearranging the left-hand side, and then expand by the distributive law as follows：

$$\sum_{i=1}^{n} (\bar{Q} - \bar{C})^2 + \sum_{i=1}^{n} (\Delta q_i - \Delta c_i)^2 - 2\sum_{i=1}^{n} (\bar{Q} - \bar{C})(\Delta q_i - \Delta c_i) \geq n(\bar{Q} - \bar{C})^2 \quad (13)$$

Note that $Q$ and $C$ are independent to $n$, and it was also proved that $\triangle p_i$ and $\triangle q_i$ are satisfied:

$$\sum_{i=1}^{n}(\Delta q_i - \Delta c_i) = \sum_{i=1}^{n}((\overline{Q} - \Delta q_i) - (\overline{C} - \Delta c_i)) = (n\overline{q} - \sum_{i=1}^{n}\Delta q_i) - (n\overline{c} - \sum_{i=1}^{n}\Delta c_i) = 0 \quad (14)$$

Therefore, after substituting 0 into the third term on the left-hand side, Ineq.(14) becomes：

$$n(\overline{Q} - \overline{C})^2 + \sum_{i=1}^{n}(\Delta q_i - \Delta c_i)^2 \geq n(\overline{Q} - \overline{C})^2 \quad (15)$$

Obviously, Ineq.(15) holds true, that is, *PAA* distance lower bounds the Euclidean distance .

Recall the definition in Eq.(5), ($\Delta p(t_e)$-$\Delta p(t_s)$ )=N$tan\theta$, we can obtain an inequality as follows($i$=1 is the starting point and $i$=$n$ is the ending point):

$$\begin{aligned}\sum_{i=1}^{n}(\Delta q_i - \Delta c_i)^2 &= (\Delta q_1 - \Delta c_1)^2 + \ldots + (\Delta q_n - \Delta c_n)^2 \geq (\Delta q_1 - \Delta c_1)^2 + (\Delta q_n - \Delta c_n)^2 \\ &= (\Delta q_1 - \Delta c_1)^2 + ((\Delta q_1 - \Delta c_1) + (N_q \tan\theta_q - N_c \tan\theta_c))^2 \\ &\geq 2(\Delta q_1 - \Delta c_1)^2 + (N_q \tan\theta_q - N_c \tan\theta_c)^2 \end{aligned} \quad (16)$$

Moreover, according to tangent trigonometric formula, it holds that:

$$\tan\theta_q - \tan\theta_c = (1 + \tan\theta_q \tan\theta_c)\tan(\theta_q - \theta_c) \quad (17)$$

Without loss of generality, let $N_q \geq N_c$(obviously ,both $N_q$ and $N_c$ satisfy $N_q \geq 1$), we can get:

$$(N_q \tan\theta_q - N_c \tan\theta_c)^2 \geq N_c^2(\tan\theta_q - \tan\theta_c)^2 \geq (\tan(\theta_q - \theta_c))^2 \geq (tfdist(Q,C))^2 \quad (18)$$

Particularly, $Q$ and $C$ are divided into one segment for the single-frame case, moreover, the trend features of $Q$ or $C$ can divide into no more than two different forms. Thus, the trend feature distance between $Q$ and C is 0 according to Definition 4 .Then, Combined with Ineq.(15) and Ineq.(18), we can get:

$$n(\overline{Q} - \overline{C})^2 + \sum_{i=1}^{n}(\Delta q_i - \Delta c_i)^2 \geq n(\overline{Q} - \overline{C})^2 + (tfdist(Q,C))^2 \geq n(dist(Q,C))^2 \quad (19)$$

Combining the above ineq (10) and ineq (19), we can get:

$$\sum_{i=1}^{n}(q_i - c_i)^2 \geq n(dist(q_i, \hat{c}_i))^2 + (tfdist(q,c))^2 \quad (20)$$

That is, the *TDIST* distance lower bounds the Euclidean distance on the single-frame case (i.e. $w$=1).

N frames can be obtained by applying the single-frame proof on every frame, that is:

$$\sqrt{\sum_{i=1}^{n}(q_i - c_i)^2} \geq \sqrt{\frac{n}{w}}\sqrt{\sum_{i=1}^{w}(dist(q,\hat{c})^2 + \frac{w}{n}(tfdist(q,c))^2)} \quad (21)$$

The right-hand side of the above inequality is *TDIST(P,Q)* and the left-hand side is the Euclidean distance between $P$ and $Q$. Therefore, the *TDIST* distance lower bounds the Euclidean distance.

# 5 EXPERIMENTAL VALIDATION

In this section, we will introduce the results of our experiment. First we present the data set used, the comparison methods and parameter settings. Then, we evaluate the performance of the proposed method in terms of false alarm rate, dimension reduction, and efficiency.

## 5.1 Data sets and parameter settings

We performed the experiments on 5 diverse time series data sets, which are provided by the UCR Time Series repository [24]. Some summary statistics of the data sets are given in Table 4. Each data set is divided into a training set and a testing set. The data sets contain classes ranging from 2 to 5, are of size from dozens to thousands, and have the lengths of time series varying from 96 to 470. In addition, the types of the data set are also diverse, including synthetic and real (recorded from some processes).

**Table 4 The Description of Experimental Data Set**

| NO | Data Set name | #Classes | Size of training set | Size of testing set | Length of time series | Type |
|---|---|---|---|---|---|---|
| 1 | ECG200 | 2 | 100 | 100 | 96 | Real |
| 2 | Two_Pattern | 4 | 1000 | 4000 | 128 | Synthetic |
| 3 | Beef | 5 | 30 | 30 | 470 | Real |
| 4 | Coffee | 2 | 28 | 28 | 286 | Real |
| 5 | CBF | 3 | 30 | 900 | 128 | Synthetic |

The main purpose of the TFSAX method is to improve the accuracy of symbolic representation and approximate distance metric. Therefore, we conduct our experiment on the classification task, of which the accuracy is determined by the distance measure. To compare the classification accuracy of the original SAX, Extended SAX (ESAX), SAX-TD and our TFSAX, we conduct the experiments using the 1 Nearest Neighbor (1-NN) classifiers. The main advantage is that the underlying distance metric is critical to the performance of 1-NN classifier; hence, the accuracy of the 1-NN classifier directly reflects the effectiveness of a distance measure. Furthermore, the 1-NN classifier is parameter free, allowing direct comparisons of different measures. To obtain the best accuracy for each method, we use the testing data to search for the best parameters $w$ and $\alpha$. For a given time series of length n, w and $\alpha$ are picked using the following criteria (to make the comparison fair, the criteria are the same as those in [3]):

For w, we search for the value from 2 up to $\lfloor n/2 \rfloor$, and double the value of w each time.

For α, we search for the value from 3 up to 10.

If two sets of parameter settings produce the same classification error rate, we choose the smaller ones.

## 5.2 Results

The experimental results of false positive rate (*fpr*), number of segments (*w*), number of symbols (*α*), and dimensionality reduction ratio (*ratio*) for SAX, ESAX, SAX-TD and TFSAX are shown in Table 5, where entries with the lowest *fpr* and ratio are highlighted.

*5.2.1 False Positive Rate*

False positive rate (*fpr*) refers to the error classification caused by the inconsistency between the label of the classification and the label of the original sequence during the classification process. It can be expressed as the ratio of the number of incorrectly classified samples to the number of original samples in the test data set. It is a commonly used measure of the classification and directly reflects the merits of the classification and can be formalized as follow:

$$FPR=(R-(R\wedge C)/C)\times 100\% \quad (22)$$

Where R represents the detection result of the algorithm and C represents the actual result in the original sequence. It can be seen from the definition that the smaller the fpr is, the better the algorithm is.

**Table5 The Overall Experiment Results**

| Data set | *SAX* | | | | *ESAX* | | | | *SAX-TD* | | | | *TFSAX* | | | |
|---|---|---|---|---|---|---|---|---|---|---|---|---|---|---|---|---|
| | *w* | *α* | *ratio* | *fpr* | *w* | *α* | ratio | *fpr* | *w* | *α* | ratio | *fpr* | *w* | *α* | ratio | *fpr* |
| *ECG200* | 32 | 10 | 0.33 | 0.12 | 32 | 10 | 1 | 0.1 | 16 | 5 | 0.34 | **0.09** | 8 | 5 | **0.27** | **0.09** |
| *Two_Pattern* | 32 | 10 | 0.25 | 0. 17 | 64 | 10 | 1.5 | 0.129 | 16 | 10 | 0.26 | 0.071 | 8 | 10 | **0.18** | **0.05** |
| *Beef* | 128 | 10 | 0.28 | 0.56 | 32 | 9 | **0.2** | 0.52 | 64 | 9 | 0.27 | 0.2 | 32 | 9 | **0.2** | **0.14** |
| *Coffee* | 128 | 10 | 0.45 | 0.496 | 4 | 5 | **0.04** | 0.179 | 8 | 3 | 0.06 | **0** | 8 | 3 | 0.18 | 0.12 |
| *CBF* | 32 | 10 | 0.25 | 0.104 | 64 | 10 | 1.5 | 0.138 | 4 | 10 | **0.07** | 0.11 | 4 | 10 | **0.07** | **0.08** |
| Average | - | - | 0.312 | 0.33 | - | - | 0.848 | 0.213 | - | - | 0.2 | 0.1 | - | - | **0.16** | **0.09** |

The experiment first adopts SAX, ESAX, SAX-TD and TFSAX to symbolize the original time series to obtain the corresponding training set and test set. And then, for each time series of the test data set in Table4, a 1-NN classifier is used to mine its nearest sequence in the corresponding training set using the above four distance metrics. If the query result of the test object is consistent with the original label of the sequence, it means that the sequence is classified correctly; otherwise it indicates misclassification. Finally, the average value of the classification results of all sequences in the training set is used as the experimental result of the algorithm in that data set.

The fpr of the four algorithms are shown in Table5, from which we can clearly find that TFSAX has the lowest error in the most of the data sets (4/5) because it using both mean feature and trend morphological feature to represent the sequence; As a contrast, SAX-TD gets the second lower error (2/5). Moreover, TFSAX significantly has achieved better result than other methods in terms of the average classification error of the test data set.

*5.2.2 Dimensionality reduction*

Since one major advantage of the SAX representation is its dimensionality or numerosity reduction, we shall compare the dimensionality reduction of our method with the SAX and the ESAX. The dimensionality reduction ratios are calculated using the w when the three methods achieve their smallest classification error rates on each data set. We will measure the dimensionality reduction ratios as follows:

Ratio =(Number of the reduced data points/Number of the original data points)×*100%* (23)

According to the definition of the symbolized representation, the reduction ratios of SAX, ESAX, SAX-TD, and TFSAX are $w/n$, $3w/n$, $(2w+1)/n$, and $2w/n$, respectively. The dimensionality reduction results of TFSAX competitive with SAX, ESAX and SAX-TD on the above five data sets are shown in Fig.5.

From Fig. 5 we can see that the dimensionality reduction effect of the TSFSA method and the dimensionality reduction effect of the SAX-TD method are basically similar, slightly better than the *SAX* method, while the dimension reduction effect of the *ESAX* is the worst. For each segment, though TFSAX, increases the trend factor and trend shape factor to describe the trend feature information of time series, it also reduces the number of segments needed to achieve the best similarity detection effect. From Table 5 we can see that the average dimension reduction ratio of TFSAX in the five experimental data sets is 0.16, which is basically equivalent to 0.2 of SAX-TD.

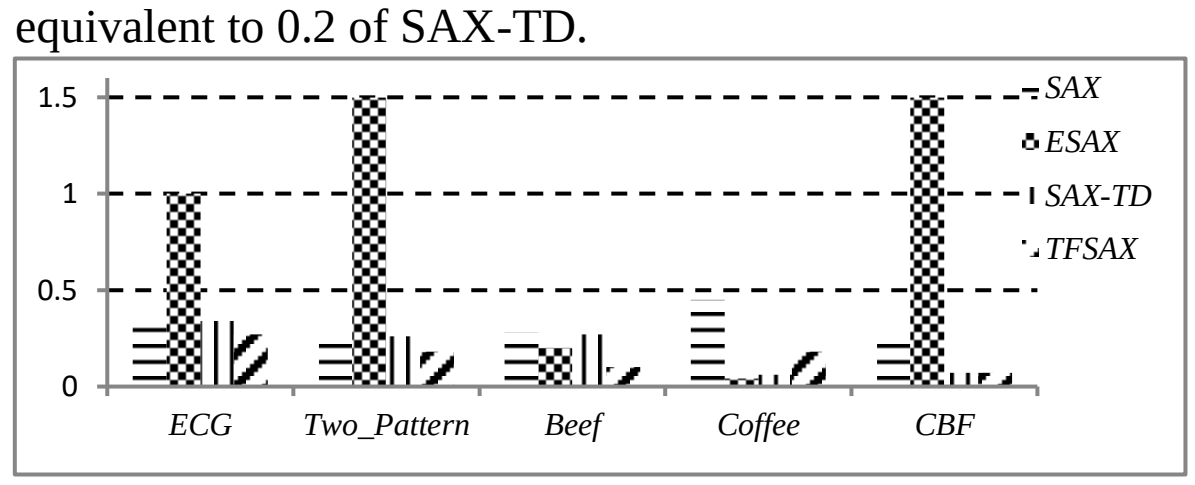

**Figure 5: Comparison of the Dimensionality reduction ratio of different methods**

*5.2.3 Lower bound*

In previous section, it has been theoretically proved that distance metric of TFSAX not only satisfies the definition of distance, but also has more compact lower boundary than SAX. This section is validated by experimentation.

The Beef dataset which containing 30 training sets and 30 test sets was used as the test data to verify the algorithm. The experiment first uses the Euclidean distance, SAX distance, SAX-TD distance and TFSAX distance to calculate the distance between any test data set and training data set (each distance measure needs to calculate 900 pairs of distances), and then calculate lower bound of SAX distance, SAX-TD distance, and TFSAX distance under condition of $w$=32, $a$=[3,4,5,6,7,8,9,10] and $a$=8, $w$=[2,4,8,16, 32,64] respectively. The average of 900 experimental results for each method was used as the final lower bound. The experimental results are shown in Fig.6 (a) and (b) respectively.

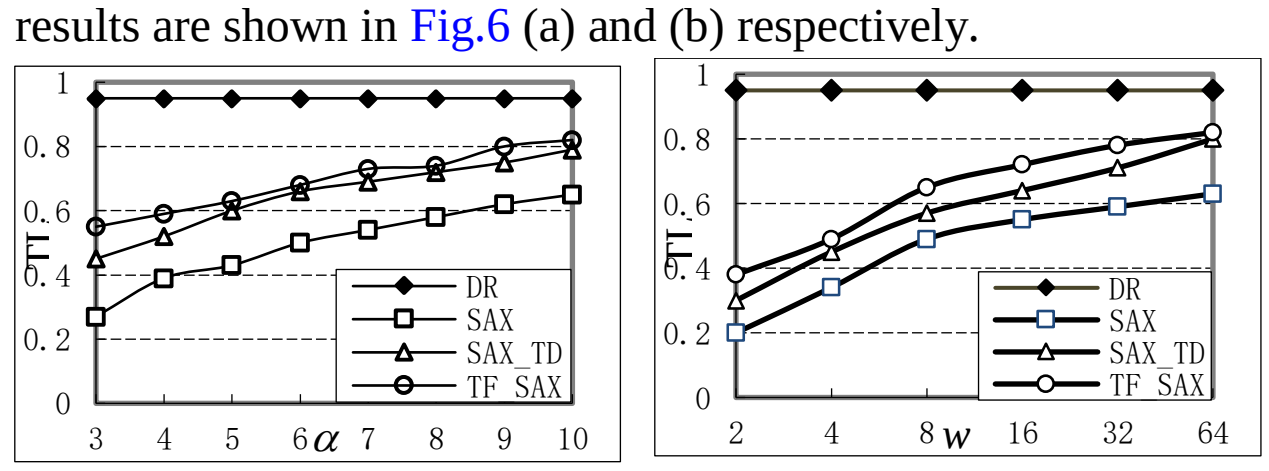

(a) *w=32, a*=[3,4,5,6,7,8,9,10]　(b) $\alpha$=8、*w*=[2,4,8,16,32,64]

**Figure 6:Comparison of the Lower Bound of Four Methods under Different Parameters**

From Fig.6 we can see that the lower bound of TFSAX distance metric of the algorithm is infinitely close to the straight line with y =1 on the vertical axis(Euclidean distance) as the number of segments and the number of symbols increase. Obviously, it not only meets the requirements of the lower bound, but also is slightly better than SAX-TD, and is always better than the original SAX. For example, the lower bound of TFSAX at $\alpha$=3 in 2.8(a) is almost close to that of SAX at $\alpha$=7, and the lower bound of TSASX at *w*=8 at 2.8(b) is almost similar to that of SAX at *w*=64.

*5.2.4 Algorithm Efficiency*

Finally, we compare the computation time of the SAX, ESAX, SAX-TD and TFSAX. The experimental environment is a machine with 4×2.9GHz processors and 12 GB RAM running 64-bit Windows Operating System. We used CBF to show the running time with different w. The maximum w values of the data sets are 2, 4, 8 16, 32 and 64 respectively. The α is fixed at the maximum value, i.e. 10. The results are shown in Fig. 7. Note that, the running time includes the transformation time (mapping values into words) and classification time (training and testing). We see that the running time increases with the increase of w. The SAX and the SAX-TD take similar amount of time while the ESAX takes more time than both especially when w becomes larger. Since the SAX-TD needs smaller parameter w for achieving the best classification accuracy in most cases, the computation time of SAX-TD is shorter than that of the SAX and ESAX in many data sets.

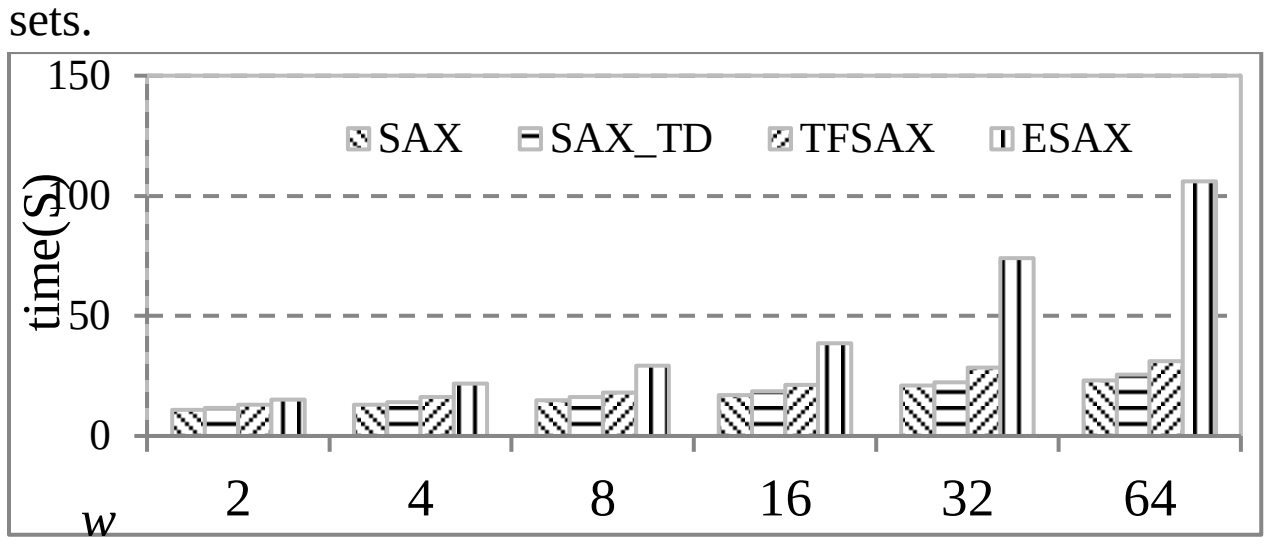

**Fig. 7. The running time of different methods on CBF data sets**

From Fig. 7, the running time of the above four algorithms will increase with the increase in the number of segments. Under the same parameters, because the TFSax algorithm needs to calculate the symbolic representation and similarity calculation of the trend morphological features, the computational time cost is higher than that of the traditional SAX algorithm and SAX_TD, but it is lower than the time consumption of the ESAX algorithm, especially When w is large. In addition, as can be seen from Figure 2.9, TFSAC increases the computational cost far less than 50% of SAX, considering that the number of segments required by TFSAC to obtain the best query result is in most cases less than or equal to the SAX required score. The number of segments and the accuracy and cost of the algorithm, the algorithm proposed in this chapter is an excellent symbolization method than SAX.

## 6 CONCLUSION

We have proposed a novel symbolic aggregate approximation representation and distance measure for time series. We firstly extract the trend feature such as trends distance factor and trend shape factor from time series to constructions trends feature triangle to quantitatively measure the different trends. And then, we design multi-resolution discretization method which uses the sequence's trend feature and mean feature to transform the trend feature into symbols. Moreover, we modify the original SAX distance measure by integrating a weighted trend distance. The new distance measure can not only keeps the important property that lower bounds the Euclidean distance, but also is tighter than that of the original SAX. According to the experimental results on diverse data sets, our improved measure decreases the classification error rate significantly and needs a smaller number of words and alphabetic symbols for achieving the best classification accuracy than the SAX and SAX-TD do. Our improved method has slightly better capability of dimensionality reduction and has similar efficiency as the SAX and SAX-TD.

For the future work, we intend to extend the method to other data mining tasks such as clustering, anomaly detection and motif discovery. The proposed method can be applied and improved in the trend feature representation method based on unequal length division.

## ACKNOWLEDGMENTS

This work has been partially supported by the CSC Scholarship and The National Key Research and Development Program of China (Nos.2018YFC0407900).